\documentclass[letterpaper]{article} 
\usepackage[submission]{aaai24}  
\usepackage{times}  
\usepackage{helvet}  
\usepackage{courier}  
\usepackage[hyphens]{url}  
\usepackage{graphicx} 
\usepackage{natbib}  
\usepackage{caption} 
\usepackage{algorithm}
\usepackage{algorithmic}

\usepackage{amssymb}
\usepackage{booktabs}
\usepackage{adjustbox}
\usepackage{multirow}

\usepackage{newfloat}
\usepackage{listings}
\DeclareCaptionStyle{ruled}{labelfont=normalfont,labelsep=colon,strut=off} 
\floatstyle{ruled}
\newfloat{listing}{tb}{lst}{}
\floatname{listing}{Listing}
\title{SegRNN: Segment Recurrent Neural Network for Long-Term Time Series Forecasting}
\author {
    Shengsheng Lin\textsuperscript{\rm 1},
    Weiwei Lin\textsuperscript{\rm 1,2},
    Wentai Wu\textsuperscript{\rm 2},
    Feiyu Zhao\textsuperscript{\rm 1},
    Ruichao Mo\textsuperscript{\rm 1},
    Haotong Zhang\textsuperscript{\rm 1}
}
\affiliations {
    \textsuperscript{\rm 1}South China University of Technology\\
    \textsuperscript{\rm 2}Peng Cheng Laboratory\\
    linss2000@foxmail.com, linww@scut.edu.cn, nnwtwu@pcl.ac.cn, zhaofy001@foxmail.com, cs\_moruichao@mail.scut.edu.cn, hoyt.zhang77@gmail.com
}

\usepackage{bibentry}

\begin{document}

\maketitle

\begin{abstract}

RNN-based methods have faced challenges in the Long-term Time Series Forecasting (LTSF) domain when dealing with excessively long look-back windows and forecast horizons. Consequently, the dominance in this domain has shifted towards Transformer, MLP, and CNN approaches. The substantial number of recurrent iterations are the fundamental reasons behind the limitations of RNNs in LTSF. To address these issues, we propose two novel strategies to reduce the number of iterations in RNNs for LTSF tasks: Segment-wise Iterations and Parallel Multi-step Forecasting (PMF). RNNs that combine these strategies, namely SegRNN, significantly reduce the required recurrent iterations for LTSF, resulting in notable improvements in forecast accuracy and inference speed. Extensive experiments demonstrate that SegRNN not only outperforms SOTA Transformer-based models  but also reduces runtime and memory usage by more than 78\%. These achievements provide strong evidence that RNNs continue to excel in LTSF tasks and encourage further exploration of this domain with more RNN-based approaches. The source code is coming soon.

\end{abstract}

\section{Introduction}
\label{introduction}

Time series forecasting involves using past observed time series data to predict future unknown time series. It finds applications in various fields, such as energy and smart grids, traffic flow control, server energy optimization \cite{energy_forecasting}. Recurrent Neural Networks (RNNs) \cite{rnn_survey}, as a deep learning architecture, have been extensively adopted for conventional time series forecasting due to their effectiveness in capturing sequential dependencies \cite{deep_forecast_survey}.

\begin{figure}[t]
  \centering
  \includegraphics[width=0.95\columnwidth]{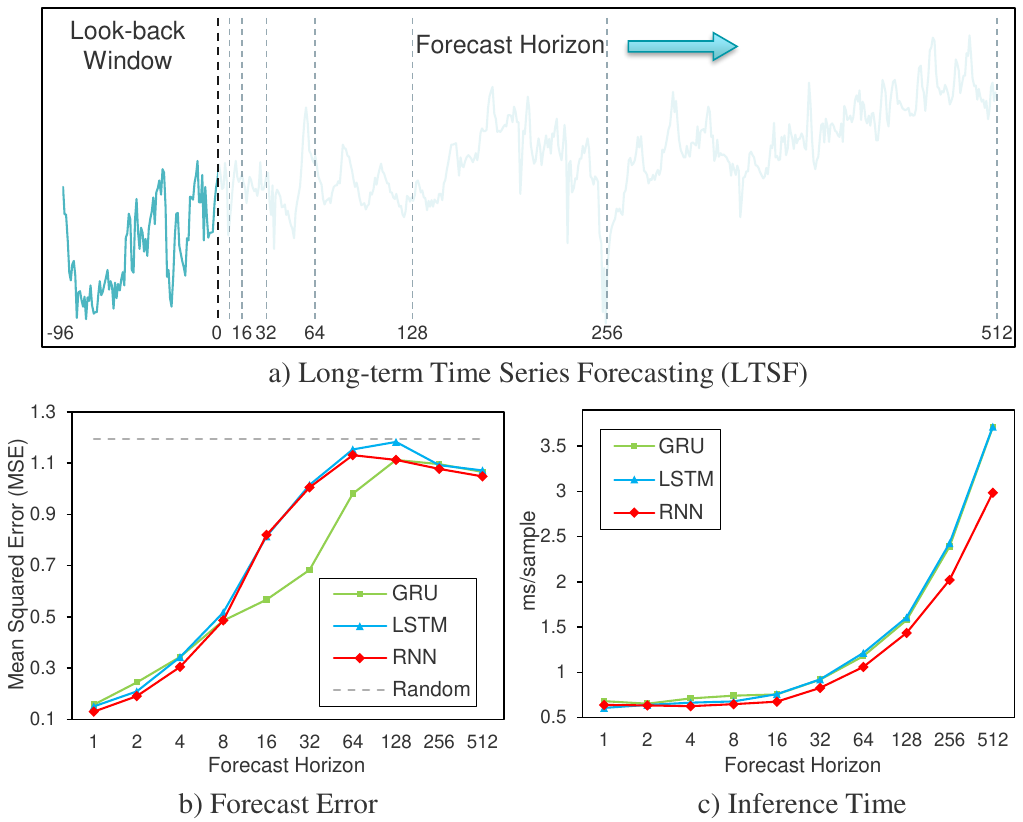}
  \caption{Challenges faced by vanilla RNN and its variants in LTSF. Data is obtained from the ETTh1 dataset.}
  \label{fig1}
\end{figure}

In recent years, there has been a shift in focus towards predicting longer horizons, known as Long-term Time Series Forecasting (LTSF) \cite{informer}. Figure \ref{fig1} (a) illustrates the concept of LTSF, where the objective is to provide richer semantic information by predicting a longer future sequence, thus offering more practical guidance. However, extending the forecast horizon poses significant challenges: (i) Forecasting further into the future leads to increased uncertainty, resulting in decreased forecast accuracy. (ii) Longer forecast horizons require models to consider a more extensive historical context for accurate predictions, significantly increasing the complexity of modeling.

While RNNs have exhibited remarkable performance in conventional time-series tasks, they have gradually lost prominence in the LTSF domain. Figure \ref{fig1} (b) and (c) illustrate the limitations of RNNs (either vanilla RNN or its variants: Long Short-Term Memory (LSTM) \cite{lstm} and Gated Recurrent Unit (GRU) \cite{gru}) in LTSF: (i) The model's forecast error rapidly increases as the forecast horizon expands, especially when the horizon reaches 64, at which point it becomes comparable to random forecasting. (ii) The models' inference time rapidly increases with the length of horizon. These observations widely support the belief that RNNs are no longer suitable for LTSF tasks that involve modeling long-term dependencies \cite{informer, fedformer}. Consequently, there is currently no prominent RNN-based solution in the LTSF field.

In contrast, Transformers \cite{transformer}, an advanced neural network architecture designed to model long-term dependencies in sequences, have achieved remarkable success in natural language processing, computer vision, and other fields. Consequently, there has been a surge in Transformer-based LTSF solutions, breaking several state-of-the-art (SOTA) records \cite{informer, fedformer, patchtst}. The undeniable efficacy of Transformers notwithstanding, their intricate design and substantial computational requirements have constrained their accessibility. Recently, there has been significant debate regarding whether the self-attention mechanism in Transformers is suitable for modeling time-series tasks \cite{dlinear, Mts-mixers}. This leads us to contemplate: \textit{Are RNNs, which are conceptually simple and structurally well-suited for modeling time-series data, truly unsuitable for LTSF tasks?}

\begin{figure}[tb]
  \centering
  \includegraphics[width=0.9\columnwidth]{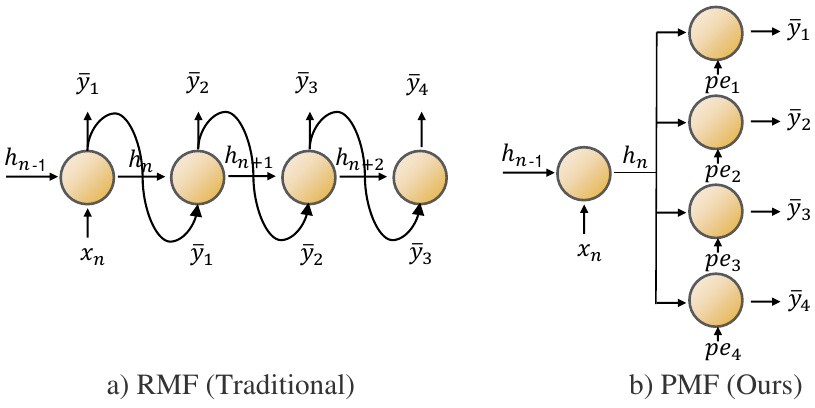}
  \caption{Comparison of Recurrent Multi-step Forecasting (RMF) and Parallel Multi-step Forecasting (PMF). The positional embedding \(pe_i\) in PMF serves as a replacement for the sequential information of the recurrent structure.}
  \label{fig2}
\end{figure}

\textit{The answer might be No.} It is well-known that RNNs suffer from the vanishing/exploding gradient problem \cite{gradient}, which limits the length of sequences they can effectively model. We can hypothesize that the current RNNs' failure in LTSF stems from excessively long look-back and forecast horizons, which result in prohibitively high recurrent iteration counts. To address this, we propose a straightforward yet powerful strategy: \textbf{Minimizing the count of recurrent iterations in RNNs while striving to retain sequential information.} Specifically, we introduce SegRNN, which is designed with two key components:

\begin{enumerate}
    \item The incorporation of segment technology in RNNs, replacing point-wise iterations with segment-wise iterations, significantly reducing the number of recurrent iterations.
    \item The introduction of the Parallel Multi-step Forecasting (PMF) strategy, further reducing the number of recurrent iterations. The comparison between PMF and the traditional Recurrent Multi-step Forecasting (RMF) is illustrated in Figure \ref{fig2}.
\end{enumerate}

Experimental results on popular LTSF benchmarks demonstrate that these two key design components significantly improve RNNs' performance in the LTSF domain. Reducing the number of recurrent iterations not only greatly improves the prediction accuracy of RNNs but also significantly enhances their inference speed. In most scenarios, SegRNN outperforms SOTA Transformer-based models, featuring a runtime and memory reduction of over 78\%. We provide strong evidence that RNNs still possess powerful capabilities in the LTSF domain.

In summary, our contributions are as follows:

\begin{itemize}
    \item We propose SegRNN, which utilizes time-series segment technique to replace point-wise iterations with segment-wise iterations in LTSF.
    \item We further introduce the PMF technique to enhance the inference speed and performance of RNNs.
    \item The proposed SegRNN outperforms the current SOTA methods while significantly reducing runtime and memory usage.
    \item The success of SegRNN demonstrates substantial improvements over existing RNN methods, highlighting the strong potential of RNNs in the LTSF domain.
\end{itemize}

\section{Related Work}
\label{related_work}

Significant efforts have been devoted to advancing the field of time series forecasting \cite{multi_forecast_servey}. With the evolution of hardware capabilities, deep learning approaches have gained prominence in uncovering patterns within time series data \cite{deep_forecast_review}. These approaches can be broadly categorized as follows:

\paragraph{Transformers.}
Originally designed for natural language processing tasks \cite{transformer}, Transformers have achieved remarkable success in various domains \cite{trans_cv_survey, trans_video_survey}. The self-attention mechanism in Transformers enables them to capture long-term dependencies in time series data, leading to a considerable body of work focused on adapting Transformers to LTSF tasks, demonstrating impressive performance \cite{trans_time_survey}. Earlier efforts, such as LogTrans \cite{LogTrans}, Informer \cite{informer}, Pyraformer \cite{pyraformer}, Autoformer \cite{Autoformer}, and FEDformer \cite{fedformer}, aimed at reducing the complexity of Transformers. More recently, PatchTST \cite{patchtst} and Crossformer \cite{crossformer} leveraged patch-based techniques from computer vision \cite{vit, MAE}, further enhancing the performance of Transformers. The patch technique, which inspired the segment-wise iterations technique in this paper, has proven to be influential.

\paragraph{MLPs.}
Multi-Layer Perceptrons (MLPs) have also found extensive use in time series forecasting \cite{NBeatsX, DEPTS, N-HiTS}. Recently, DLinear achieved superiority over then-state-of-the-art Transformer-based models through a simple linear layer and channel-independent strategy \cite{dlinear}. The success of DLinear has spurred the development of a plethora of MLPs in LTSF, including MTS-Mixers \cite{Mts-mixers}, TSMixer \cite{TSMixer}, and TiDE \cite{tide}. The accomplishments of these MLP-based models have raised questions about the necessity of employing complex and cumbersome Transformers for time series prediction. 

\paragraph{CNNs.}
Initially applied in image processing for capturing local patterns and extracting meaningful features \cite{AlexNet, ResNet}, Convolutional Neural Networks (CNNs) have also shown remarkable performance in the time series domain \cite{TCN, TCN2, MLCNN}. Recently, CNN-based models such as MICN \cite{micn}, TimesNet \cite{wu2023timesnet}, and SCINet \cite{SCInet} have demonstrated impressive results in the LTSF field. 

\paragraph{RNNs.}
Recurrent Neural Networks (RNNs) have long been the primary choice for time series forecasting tasks due to their ability to handle sequential data. Numerous efforts have been devoted to utilizing RNNs for short-term and probabilistic forecasting, achieving significant advancements \cite{LSTnet, C2FAR, MQRNN, RNN-Adap}. However, in the LTSF domain with excessively long look-back windows and forecast horizons, RNNs have been considered inadequate for effectively capturing long-term dependencies, leading to their gradual abandonment \cite{informer, fedformer}. The emergence of SegRNN aims to challenge and change this situation by attempting to address these limitations.

\section{Preliminaries}
\label{preliminaries}

This section introduces the formulation of the LTSF problem, the Channel Independent (CI) strategy, and the fundamental RNN and its variants.

\subsection{LTSF Problem Formulation}
The LTSF problem deals with predicting the future time series \(Y \in \mathbb{R}^{H \times C}\) based on a historical multivariate time series (MTS) \(X \in \mathbb{R}^{L \times C}\). Here, \(L\) represents the length of the historical look-back window, \(C\) denotes the number of feature dimensions or channels, and \(H\) signifies the length of the forecast horizon. The goal of the LTSF task is to extend the forecasting horizon \(H\) to its maximum potential (e.g., up to 720), which poses a considerable challenge.

\subsection{Channel Independent Strategy}
Intuitively, it may appear optimal to use all historical variables in a MTS to forecast all future variables simultaneously, as it captures the interrelationships between the variables. However, recent studies have shown that the Channel Independent (CI) strategy surpasses this traditional approach \cite{CICD}. The CI strategy aims to identify a function \(f: X^{(i)} \in \mathbb{R}^{L} \rightarrow Y^{(i)} \in \mathbb{R}^{H}\) that maps the univariate historical time series data to the future time series values, as opposed to \(f: X \in \mathbb{R}^{L \times C} \rightarrow Y \in \mathbb{R}^{H \times C}\), which maps the multivariate historical time series data to the future time series values. The current SOTA LTSF models have embraced the CI strategy \cite{dlinear, patchtst, tide}, and we similarly align ourselves with this approach. Furthermore, our model introduces a channel identifier \cite{STID} to enhance its predictive capability for multivariate sequences, which will be discussed comprehensively in the following sections.

\subsection{RNN Variants}
\begin{figure}[htb]
  \centering
  \includegraphics[width=0.95\columnwidth]{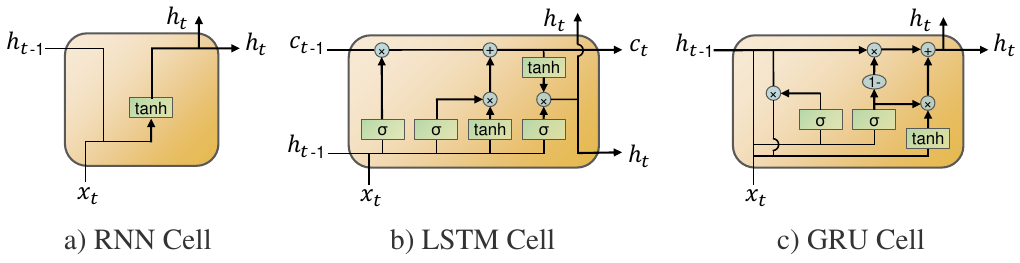}
  \caption{Comparison of the cell of RNN and its variants.}
  \label{fig3}
\end{figure}

The vanilla RNN faces challenges such as vanishing and exploding gradients, which hinder the model's convergence during training \cite{gradient}. To address these issues, LSTM and GRU architectures optimize the structure of their cell units. Figure \ref{fig3} illustrates the differences in cell unit structures among these fundamental RNN variants. For more comprehensive information, we strongly recommend referring to the original papers \cite{lstm, gru}.

The proposed SegRNN is not limited to a specific RNN cell structure. Considering the stable performance of GRU in practical scenarios, this paper employs GRU as an exemplar to illustrate the architecture and performance of SegRNN. Therefore, for consistency throughout the text, the SegRNN model is assumed to be based on the GRU cell unless explicitly stated otherwise.

\section{Model Architecture}
\label{model_architecture}

\begin{figure*}[bt]
  \centering
  \includegraphics[width=0.75\textwidth]{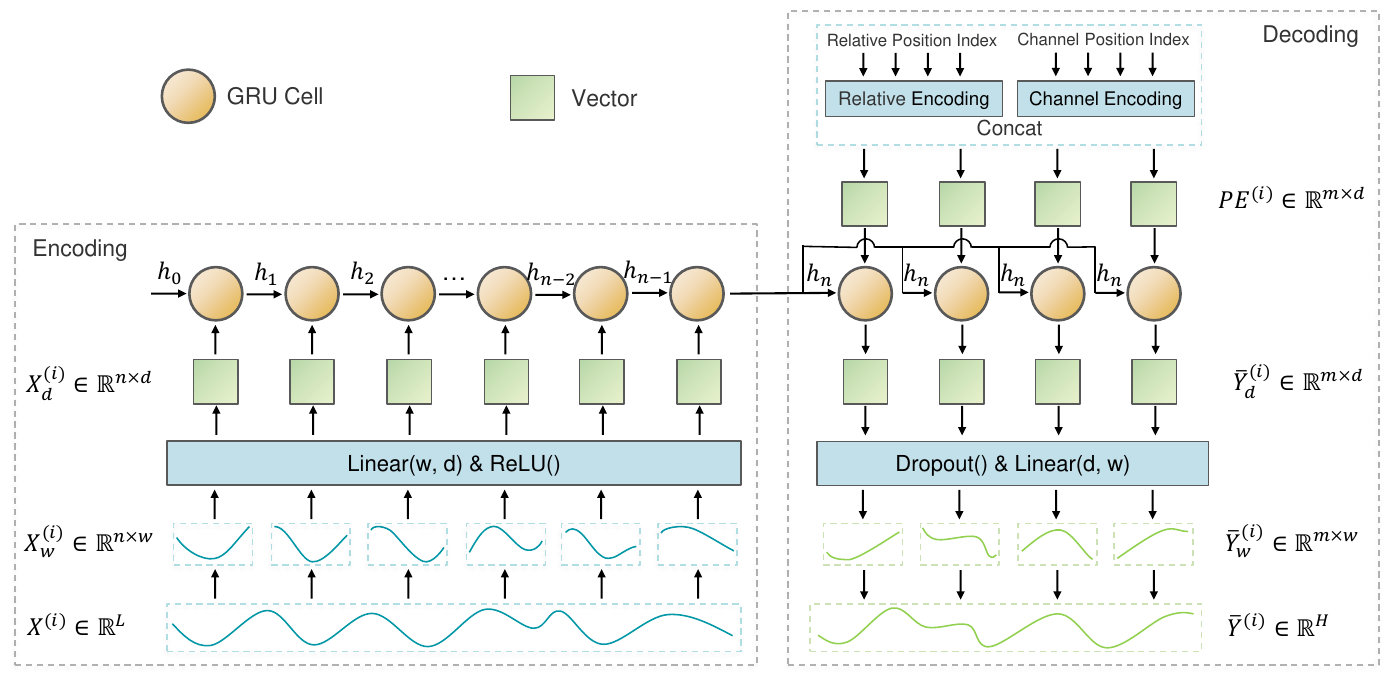}
  \caption{The model architecture of SegRNN.}
  \label{fig4}
\end{figure*}

The recurrent iterative nature of RNNs poses challenges for effective convergence when modeling extensive long sequences. SegRNN aims to reduce the number of recurrent iterations to facilitate its convergence. Specifically, SegRNN employs the following strategies: 
\begin{enumerate}
    \item In the encoding phase, it replaces the original time point-wise iterations with sequence segment-wise iterations, effectively reducing the number of iterations from \(L\) to \(L/w\).
    \item In the decoding phase, it utilizes the PMF strategy to further reduce the number of iterations from \(H/w\) to 1.
\end{enumerate}
The substantial reduction in the number of recurrent iterations in RNN not only leads to a remarkable performance improvement but also results in a significant increase in inference speed. The model architecture of SegRNN is illustrated in Figure \ref{fig4}.

\subsection{Encoding}
\paragraph{Segment partition and projection.}
Given a sequence channel \(X^{(i)} \in \mathbb{R}^L\), it can be partitioned into segments \(X^{(i)}_w \in \mathbb{R}^{n \times w}\), where \(w\) represents the window length of each segment, and \(n=\frac{L}{w}\) denotes the number of segments. These segments, \(X^{(i)}_w \in \mathbb{R}^{n \times w}\), are then transformed to \(X^{(i)}_d \in \mathbb{R}^{n \times d}\) through a learnable linear projection \(W_{prj} \in \mathbb{R}^{w \times d}\) followed by a ReLU activation, where \(d\) represents the dimensionality of the hidden state of the GRU.

\paragraph{Recursive encoding.}
Subsequently, the transformed \(X^{(i)}_d\) is fed into the GRU for recurrent iterations to capture temporal features. Specifically, for \(x_t \in \mathbb{R}^d \) in \(X^{(i)}_d\), the entire process within the GRU cell can be formulated as:
\[z_t  = \sigma \left ( W_z \cdot \left [ h_{t-1}, x_t \right ] \right ),\]
\[r_t  = \sigma \left ( W_r \cdot \left [ h_{t-1}, x_t \right ] \right ),\]
\[\tilde{h}_t  = \tanh \left ( W \cdot \left [ r_t \times h_{t-1}, x_t \right ] \right ),\]
\[h_t  = \left ( 1-z_t \right ) \times h_{t-1} + z_t \times \tilde{h}_t. \]

After \(n\) recurrent iterations, the hidden feature \(h_n\) obtained from the last step already encapsulates all the temporal features of the original sequence \(X^{(i)}\). This hidden feature will be passed to the Decoding part for the subsequent steps of inference and prediction.

\subsection{Decoding}
RMF is a straightforward method for accomplishing multi-step prediction. In RMF, single-step predictions are made to obtain \(\bar{y}_t\). The predicted \(\bar{y}_t\) is then used as input for the subsequent prediction of \(\bar{y}_{t+1}\), and this process continues until the complete set of prediction results is obtained. By incorporating the segment technique from the Encoding phase with RMF, the number of iterations required for a prediction horizon of \(H\) is reduced to \(H/w\).

However, despite the reduction in the number of recurrent iterations, recursive decoding has its limitations. These include: (i) the accumulation of errors resulting from recursive predictions, and (ii) the sequential nature of recursion, which hampers parallel computation within training examples and restricts improvements in inference speed \cite{transformer}. To address these limitations, we propose a novel prediction strategy called Parallel Multi-step Forecasting (PMF), as described below.

\paragraph{Positional embeddings.}

During the decoding phase, the sequential order between segments is lost due to the break in the recurrent recursion. To address this, \(m\) corresponding positional embeddings, denoted as \(PE^{(i)} \in \mathbb{R}^{m \times d}\), are generated to identify the positions of the segments. Here, \(m = \frac{H}{w}\) represents the number of windows obtained by partitioning the prediction horizon into segments. Each positional embedding \(pe^{(i)} \in \mathbb{R}^{d}\) is constructed by concatenating the relative position encoding \(rp \in \mathbb{R}^{\frac{d}{2}}\) and the channel position encoding \(cp \in \mathbb{R}^{\frac{d}{2}}\).

\begin{figure}[ht]
  \centering
  \includegraphics[width=0.9\columnwidth]{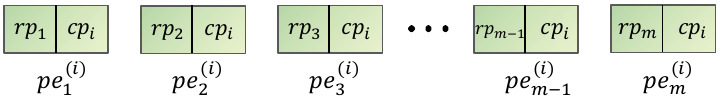}
  \caption{Positional embeddings \(PE^{(i)}\) for the target sequence \(Y^{(i)}\).}
  \label{fig5}
\end{figure}

Figure \ref{fig5} illustrates the positional embeddings for the target sequence \(Y^{(i)}\). The relative position encoding indicates the position of each segment that needs to be predicted within the complete sequence \(Y^{(i)}\). Meanwhile, the channel position encoding represents the channel position \(i \in 1, 2, ..., C\) of the current sequence \(Y^{(i)}\) within the multi-channel sequence \(Y \in \mathbb{R}^{H \times C}\) \cite{STID}. The inclusion of the channel position encoding partially compensates for the limitations of the CI strategy in capturing relationships between variables, thereby enhancing the model's performance.

\paragraph{Parallel decoding.}
In the Decoding phase, the same GRU cell used in the Encoding phase is shared. Specifically, the final state \(h_n\) obtained from the Encoding phase is duplicated \(m\) times and combined with the \(m\) positional embeddings from \(PE^{(i)}\). These pairs are then simultaneously processed in parallel by the GRU cell. This parallel processing generates \(m\) output vectors, each with a length of \(d\), denoted as \(\bar{Y}^{(i)}_d \in \mathbb{R}^{m \times d}\). It is important to note that this approach differs from the previous recursive processing in RMF, as the computation of each vector is independent of the previous time step's result. As a result, intra-sample parallel computation is achieved, leading to improved inference speed. Additionally, prediction errors do not accumulate with the number of iterations, resulting in enhanced prediction accuracy.

\paragraph{Prediction and sequence recovery.}
The \(\bar{Y}^{(i)}_d\) undergoes a Dropout layer, randomly dropping out a certain proportion of values for regularization purposes. Subsequently, it is transformed into \(\bar{Y}^{(i)}_w \in \mathbb{R}^{m \times w}\) using a learnable linear prediction layer \(W_{prd} \in \mathbb{R}^{d \times w}\). Furthermore, \(\bar{Y}^{(i)}_w\) is reshaped into \(\bar{Y}^{(i)} \in \mathbb{R}^H\), representing the final prediction result.

\subsection{Normalization and Evaluation}
\paragraph{Instance normalization.} Time series data often experience distribution shift issues, and employing simple sample normalization strategies can help alleviate this problem. In this paper, we utilize a simple sample normalization strategy \cite{dlinear} that involves subtracting the last value of the sequence from the input bofore encoding and subsequently adding back the value after decoding, formulated as:
\[ x^{(i)}_{1:L} = x^{(i)}_{1:L} - x^{(i)}_{L},\]
\[ \bar{y}^{(i)}_{1:L} = \bar{y}^{(i)}_{1:L} + x^{(i)}_{L}.\]

\paragraph{Loss function.} The mean absolute error (MAE) is employed as the loss function for our model, defined as:
\[\mathcal{L}(Y, \bar{Y}) = \frac{1}{HC}\sum_{t=1}^{H}\sum_{i=1}^{C}|\bar{y}^{(i)}_t - y^{(i)}_t|.\]

\section{Experiments}
In this section, we present the main experimental results on popular LTSF benchmarks. Furthermore, we conduct an ablation study to analyze the impact of the segment-wise iterations and the PMF strategies on the effectiveness of RNN in LTSF. We also investigate the influence of some important parameters on SegRNN. 
All experiments in this section are implemented in PyTorch and executed on two NVIDIA T4 GPUs, each equipped with 16GB of memory.

\subsection{Experimental Setup}
An overview of the experimental setup is presented here, and for further details, please refer to the Appendix.
\paragraph{Datasets.}
The performance evaluation of SegRNN is carried out on 7 popular datasets in the LTSF domain, comprising 4 ETT datasets (ETTh1, ETTh2, ETTm1, ETTm2), as well as Traffic, Electricity, and Weather datasets. The statistics of these datasets are presented in Table \ref{dataset}.
\begin{table}[htb]
\centering
\begin{adjustbox}{max width=\columnwidth}
\begin{tabular}{@{}lccccccc@{}}
\toprule
Datasets & ETTh1 & ETTh2 & ETTm1 & ETTm2 & Electricity & Traffic & Weather \\ \midrule
Channels & 7 & 7 & 7 & 7 & 321 & 862 & 21 \\
Frequency & 1 hour & 1 hour & 15 mins & 15 mins & 1 hour & 1 hour & 10 mins \\
Timesteps & 17,420 & 17,420 & 69,680 & 69,680 & 26,304 & 17,544 & 52,696 \\ \bottomrule
\end{tabular}
\end{adjustbox}
\caption{Summary of datasets for evaluation.}
\label{dataset}
\end{table}

\paragraph{Baselines and metrics.}
As baselines, we select SOTA and representative models in the LTSF domain, including the following categories: (i) Transformers: PatchTST \cite{patchtst}, FEDformer \cite{fedformer}, Informer \cite{informer}; (ii) MLPs: TiDE \cite{tide}, Dlinear \cite{dlinear}; (iii) CNNs: MICN \cite{micn}, TimesNet \cite{wu2023timesnet}; (iv) RNNs: DeepAR \cite{deepar}, GRU \cite{gru}. It is worth mentioning that while DeepAR and GRU were not initially designed for LTSF, we include them as baselines due to the limited presence of prominent RNN solutions in the field. The proposed SegRNN aims to fill this gap.

Two commonly employed evaluation metrics, Mean Squared Error (MSE) and Mean Absolute Error (MAE), are used here to assess the performance of the LTSF models.

\paragraph{Model configuration.}
The uniform configuration of SegRNN consists of a look-back of 720, a segment length of 48, a single GRU layer, a hidden size of 512, 30 training epochs, a learning rate decay of 0.8 after the initial 3 epochs, and early stopping with a patience of 10. The dropout rate, batch size, and learning rate vary based on the scale of the data. 

\subsection{Main Result}
\begin{table*}[t]
\centering
\begin{adjustbox}{max width=\textwidth}
\begin{tabular}{@{}cccccccc|cccccc|cccc|cccc@{}}
\toprule
\multicolumn{2}{c|}{Categories}  & \multicolumn{6}{c|}{RNNs} & \multicolumn{6}{c|}{Transformers}   & \multicolumn{4}{c|}{MLPs}   & \multicolumn{4}{c}{CNNs} \\ \midrule
\multicolumn{2}{c|}{Models}   & \multicolumn{2}{c}{\begin{tabular}[c]{@{}c@{}}SegRNN\\ (ours)\end{tabular}} & \multicolumn{2}{c}{\begin{tabular}[c]{@{}c@{}}DeepAR\\ (2020)\end{tabular}} & \multicolumn{2}{c|}{\begin{tabular}[c]{@{}c@{}}GRU\\ (2014)\end{tabular}} & \multicolumn{2}{c}{\begin{tabular}[c]{@{}c@{}}PatchTST\\ (2023)\end{tabular}} & \multicolumn{2}{c}{\begin{tabular}[c]{@{}c@{}}FEDformer\\ (2022)\end{tabular}} & \multicolumn{2}{c|}{\begin{tabular}[c]{@{}c@{}}Informer\\ (2021)\end{tabular}} & \multicolumn{2}{c}{\begin{tabular}[c]{@{}c@{}}TiDE\\ (2023)\end{tabular}} & \multicolumn{2}{c|}{\begin{tabular}[c]{@{}c@{}}Dlinear\\ (2023)\end{tabular}} & \multicolumn{2}{c}{\begin{tabular}[c]{@{}c@{}}MICN\\ (2023)\end{tabular}} & \multicolumn{2}{c}{\begin{tabular}[c]{@{}c@{}}TimesNet\\ (2023)\end{tabular}} \\ \midrule
\multicolumn{2}{c|}{Metric}   & MSE & MAE & MSE & MAE & MSE   & MAE   & MSE  & MAE  & MSE   & MAE  & MSE   & MAE  & MSE   & MAE   & MSE  & MAE  & MSE   & MAE   & MSE  & MAE  \\ \midrule
\multicolumn{1}{c|}{\multirow{4}{*}{\rotatebox{90}{ETTh1}}} & \multicolumn{1}{c|}{96}  & \textbf{0.341}  & \textbf{0.376}  & 1.128  & 0.81   & 1.126 & 0.831 & \underline{0.37} & 0.4  & 0.376 & 0.415   & 0.941 & 0.769   & 0.375 & \underline{0.398} & 0.375   & 0.399   & 0.421 & 0.431 & 0.384   & 0.402   \\
\multicolumn{1}{c|}{}  & \multicolumn{1}{c|}{192} & \textbf{0.385}  & \textbf{0.402}  & 1.077  & 0.776  & 1.133 & 0.79  & 0.413   & 0.429   & 0.423 & 0.446   & 1.007 & 0.786   & 0.412 & 0.422 & \underline{0.405}   & \underline{0.416}   & 0.474 & 0.487 & 0.436   & 0.429   \\
\multicolumn{1}{c|}{}  & \multicolumn{1}{c|}{336} & \textbf{0.401}  & \textbf{0.417}  & 1.043  & 0.766  & 1.053 & 0.774 & \underline{0.422}   & 0.44 & 0.444 & 0.462   & 1.038 & 0.784   & 0.435 & \underline{0.433} & 0.439   & 0.443   & 0.569 & 0.551 & 0.491   & 0.469   \\
\multicolumn{1}{c|}{}  & \multicolumn{1}{c|}{720} & \textbf{0.434}  & \textbf{0.447}  & 1.075  & 0.795  & 1.077 & 0.795 & \underline{0.447}   & 0.468   & 0.469 & 0.492   & 1.144 & 0.857   & 0.454 & \underline{0.465} & 0.472   & 0.49 & 0.77  & 0.672 & 0.521   & 0.5  \\ \midrule
\multicolumn{1}{c|}{\multirow{4}{*}{\rotatebox{90}{ETTh2}}} & \multicolumn{1}{c|}{96}  & \textbf{0.263}  & \textbf{0.32}   & 2.602  & 1.329  & 2.1   & 1.094 & 0.274   & 0.337   & 0.332 & 0.374   & 1.549 & 0.952   & \underline{0.27}  & \underline{0.336} & 0.289   & 0.353   & 0.299 & 0.364 & 0.34 & 0.374   \\
\multicolumn{1}{c|}{}  & \multicolumn{1}{c|}{192} & \textbf{0.321}  & \textbf{0.36}   & 3.198  & 1.368  & 2.548 & 1.249 & 0.341   & 0.382   & 0.407 & 0.446   & 3.792 & 1.542   & \underline{0.332} & \underline{0.38}  & 0.383   & 0.418   & 0.441 & 0.454 & 0.402   & 0.414   \\
\multicolumn{1}{c|}{}  & \multicolumn{1}{c|}{336} & \textbf{0.325}  & \textbf{0.374}  & 3.139  & 1.353  & 3.142 & 1.343 & \underline{0.329}   & \underline{0.384}   & 0.4   & 0.447   & 4.215 & 1.642   & 0.36  & 0.407 & 0.448   & 0.465   & 0.654 & 0.567 & 0.452   & 0.452   \\
\multicolumn{1}{c|}{}  & \multicolumn{1}{c|}{720} & \underline{0.394}  & \underline{0.424}  & 3.134  & 1.352  & 3.138 & 1.343 & \textbf{0.379}   & \textbf{0.422}   & 0.412 & 0.469   & 3.656 & 1.619   & 0.419 & 0.451 & 0.605   & 0.551   & 0.956 & 0.716 & 0.462   & 0.468   \\ \midrule
\multicolumn{1}{c|}{\multirow{4}{*}{\rotatebox{90}{ETTm1}}} & \multicolumn{1}{c|}{96}  & \textbf{0.282}  & \textbf{0.335}  & 1.283  & 0.857  & 0.706 & 0.587 & \underline{0.293}   & 0.346   & 0.326 & 0.39 & 0.626 & 0.56 & 0.306 & 0.349 & 0.299   & \underline{0.343}   & 0.316 & 0.362 & 0.338   & 0.375   \\
\multicolumn{1}{c|}{}  & \multicolumn{1}{c|}{192} & \textbf{0.319}  & \textbf{0.36}   & 1.181  & 0.839  & 0.946 & 0.738 & \underline{0.333}   & 0.37 & 0.365 & 0.415   & 0.725 & 0.619   & 0.335 & 0.366 & 0.335   & \underline{0.365}   & 0.363 & 0.39  & 0.374   & 0.387   \\
\multicolumn{1}{c|}{}  & \multicolumn{1}{c|}{336} & \textbf{0.349}  & \textbf{0.383}  & 1.249  & 0.846  & 1.048 & 0.767 & 0.369   & 0.392   & 0.392 & 0.425   & 1.005 & 0.741   & \underline{0.364} & \underline{0.384} & 0.369   & 0.386   & 0.408 & 0.426 & 0.41 & 0.411   \\
\multicolumn{1}{c|}{}  & \multicolumn{1}{c|}{720} & \textbf{0.41}   & \underline{0.418}  & 1.075  & 0.77   & 1.076 & 0.765 & 0.416   & 0.42 & 0.446 & 0.458   & 1.133 & 0.845   & \underline{0.413} & \textbf{0.413} & 0.425   & 0.421   & 0.481 & 0.476 & 0.478   & 0.45 \\ \midrule
\multicolumn{1}{c|}{\multirow{4}{*}{\rotatebox{90}{ETTm2}}} & \multicolumn{1}{c|}{96}  & \textbf{0.158}  & \textbf{0.241}  & 3.418  & 1.581  & 0.487 & 0.518 & 0.166   & 0.256   & 0.18  & 0.271   & 0.355 & 0.462   & \underline{0.161} & \underline{0.251} & 0.167   & 0.26 & 0.179 & 0.275 & 0.187   & 0.267   \\
\multicolumn{1}{c|}{}  & \multicolumn{1}{c|}{192} & \textbf{0.215}  & \textbf{0.283}  & 3.894  & 1.635  & 1.725 & 0.984 & \underline{0.223}   & 0.296   & 0.252 & 0.318   & 0.595 & 0.586   & \textbf{0.215} & \underline{0.289} & 0.224   & 0.303   & 0.307 & 0.376 & 0.249   & 0.309   \\
\multicolumn{1}{c|}{}  & \multicolumn{1}{c|}{336} & \textbf{0.263}  & \textbf{0.317}  & 3.247  & 1.377  & 2.091 & 1.082 & 0.274   & 0.329   & 0.324 & 0.364   & 1.27  & 0.871   & \underline{0.267} & \underline{0.326} & 0.281   & 0.342   & 0.325 & 0.388 & 0.321   & 0.351   \\
\multicolumn{1}{c|}{}  & \multicolumn{1}{c|}{720} & \textbf{0.33}   & \textbf{0.366}  & 2.588  & 1.313  & 2.301 & 1.144 & 0.362   & 0.385   & 0.41  & 0.42 & 3.001 & 1.267   & \underline{0.352} & \underline{0.383} & 0.397   & 0.421   & 0.502 & 0.49  & 0.408   & 0.403   \\ \midrule
\multicolumn{1}{c|}{\multirow{4}{*}{\rotatebox{90}{Electricity}}} & \multicolumn{1}{c|}{96}  & \textbf{0.128}  & \textbf{0.219}  & 0.575  & 0.535  & 0.404 & 0.434 & \underline{0.129}   & \underline{0.222}   & 0.186 & 0.302   & 0.304 & 0.393   & 0.132 & 0.229 & 0.14 & 0.237   & 0.164 & 0.269 & 0.168   & 0.272   \\
\multicolumn{1}{c|}{}  & \multicolumn{1}{c|}{192} & \underline{0.148}  & \textbf{0.239}  & 1.061  & 0.809  & 0.844 & 0.717 & \textbf{0.147}   & \underline{0.24} & 0.197 & 0.311   & 0.327 & 0.417   & \textbf{0.147} & 0.243 & 0.153   & 0.249   & 0.177 & 0.285 & 0.184   & 0.289   \\
\multicolumn{1}{c|}{}  & \multicolumn{1}{c|}{336} & 0.166  & \textbf{0.258}  & 1.04   & 0.795  & 1.015 & 0.775 & \underline{0.163}   & \underline{0.259}   & 0.213 & 0.328   & 0.333 & 0.422   & \textbf{0.161} & 0.261 & 0.169   & 0.267   & 0.193 & 0.304 & 0.198   & 0.3  \\
\multicolumn{1}{c|}{}  & \multicolumn{1}{c|}{720} & 0.201  & \textbf{0.29}   & 1.048  & 0.804  & 1.041 & 0.783 & \underline{0.197}   & \textbf{0.29} & 0.233 & 0.344   & 0.351 & 0.427   & \textbf{0.196} & \underline{0.294} & 0.203   & 0.301   & 0.212 & 0.321 & 0.22 & 0.32 \\ \midrule
\multicolumn{1}{c|}{\multirow{4}{*}{\rotatebox{90}{Traffic}}}  & \multicolumn{1}{c|}{96}  & 0.543  & \textbf{0.235}  & 1.377  & 0.717  & 1.349 & 0.715 & \underline{0.36} & \underline{0.249}   & 0.576 & 0.359   & 0.733 & 0.41 & \textbf{0.336} & 0.253 & 0.41 & 0.282   & 0.519 & 0.309 & 0.593   & 0.321   \\
\multicolumn{1}{c|}{}  & \multicolumn{1}{c|}{192} & 0.567  & \textbf{0.246}  & 1.442  & 0.75   & 1.42  & 0.75  & \underline{0.379}   & \underline{0.256}   & 0.61  & 0.38 & 0.777 & 0.435   & \textbf{0.346} & 0.257 & 0.423   & 0.287   & 0.537 & 0.315 & 0.617   & 0.336   \\
\multicolumn{1}{c|}{}  & \multicolumn{1}{c|}{336} & 0.602  & \textbf{0.256}  & 1.489  & 0.778  & 1.47  & 0.776 & \underline{0.392}   & 0.264   & 0.608 & 0.375   & 0.776 & 0.434   & \textbf{0.355} & \underline{0.26}  & 0.436   & 0.296   & 0.534 & 0.313 & 0.629   & 0.336   \\
\multicolumn{1}{c|}{}  & \multicolumn{1}{c|}{720} & 0.671  & \underline{0.281}  & 1.526  & 0.793  & 1.524 & 0.794 & \underline{0.432}   & 0.286   & 0.621 & 0.375   & 0.827 & 0.466   & \textbf{0.386} & \textbf{0.273} & 0.466   & 0.315   & 0.577 & 0.325 & 0.64 & 0.35 \\ \midrule
\multicolumn{1}{c|}{\multirow{4}{*}{\rotatebox{90}{Weather}}}  & \multicolumn{1}{c|}{96}  & \textbf{0.142}  & \textbf{0.181}  & 0.278  & 0.301  & 0.183 & 0.231 & \underline{0.149}   & \underline{0.198}   & 0.238 & 0.314   & 0.354 & 0.405   & 0.166 & 0.222 & 0.176   & 0.237   & 0.161 & 0.229 & 0.172   & 0.22 \\
\multicolumn{1}{c|}{}  & \multicolumn{1}{c|}{192} & \textbf{0.186}  & \textbf{0.227}  & 0.376  & 0.369  & 0.299 & 0.336 & \underline{0.194}   & \underline{0.241}   & 0.275 & 0.329   & 0.419 & 0.434   & 0.209 & 0.263 & 0.22 & 0.282   & 0.22  & 0.281 & 0.219   & 0.261   \\
\multicolumn{1}{c|}{}  & \multicolumn{1}{c|}{336} & \textbf{0.237}  & \textbf{0.269}  & 0.568  & 0.527  & 0.376 & 0.374 & \underline{0.245}   & \underline{0.282}   & 0.339 & 0.377   & 0.583 & 0.543   & 0.254 & 0.301 & 0.265   & 0.319   & 0.278 & 0.331 & 0.28 & 0.306   \\
\multicolumn{1}{c|}{}  & \multicolumn{1}{c|}{720} & \textbf{0.31}   & \textbf{0.32}   & 0.571  & 0.533  & 0.459 & 0.433 & 0.314   & \underline{0.334}   & 0.389 & 0.409   & 0.916 & 0.705   & 0.313 & 0.34  & 0.323   & 0.362   & \underline{0.311} & 0.356 & 0.365   & 0.359   \\ \midrule
\multicolumn{2}{c|}{Count}   & \multicolumn{2}{c}{\textbf{50}}   & \multicolumn{2}{c}{0} & \multicolumn{2}{c|}{0}  & \multicolumn{2}{c}{\underline{31}}  & \multicolumn{2}{c}{0} & \multicolumn{2}{c|}{0} & \multicolumn{2}{c}{29} & \multicolumn{2}{c|}{4}   & \multicolumn{2}{c}{1}  & \multicolumn{2}{c}{0}   \\ \bottomrule
\end{tabular}
\end{adjustbox}
\caption{Multivariate long-term time series forecasting results. The forecast horizon \(H \in \{96, 192, 336, 720\}\) is set for all datasets. The reported SegRNN results are averaged over 5 runs. The best results are highlighted in \textbf{bold} and the second best are \underline{underlined}. The \textit{Count} row counts the total number of times each method obtained the best or second results.} 
\label{main_result}
\end{table*}

The multivariate long-term time series forecasting results of SegRNN and other baselines are presented in Table \ref{main_result}. Remarkably, SegRNN achieved top-two positions in 50 out of 56 metrics across all scenarios, including 45 first-place rankings, signifying its significant superiority over other baselines, including the current SOTA transformer-based model, PatchTST. SegRNN demonstrated outstanding performance on the ETT and Weather datasets, nearly surpassing the SOTA performance across all metrics. In larger-scale datasets such as Electricity and Traffic, where the channel numbers exceed 300 and 800, respectively, SegRNN's performance experienced a slight decrease. This is likely due to the relatively smaller capacity of the SegRNN model, as it is built on a single GRU layer. However, even in these cases, SegRNN demonstrated competitive or superior performance compared to the competing models. 

Regarding the RNN-based methods GRU and DeepAR, SegRNN achieved MSE improvements of 75\% and 80\%, respectively. This provides strong evidence that SegRNN significantly enhances the performance of existing RNN methods in the LTSF domain. These results highlight the success of the SegRNN design and demonstrate that RNN methods still hold strong potential in the current LTSF domain.

\subsection{Ablation Studies}
\paragraph{Segment-wise iterations vs. point-wise iterations.}

\begin{figure}[htb]
  \centering
  \includegraphics[width=0.95\columnwidth]{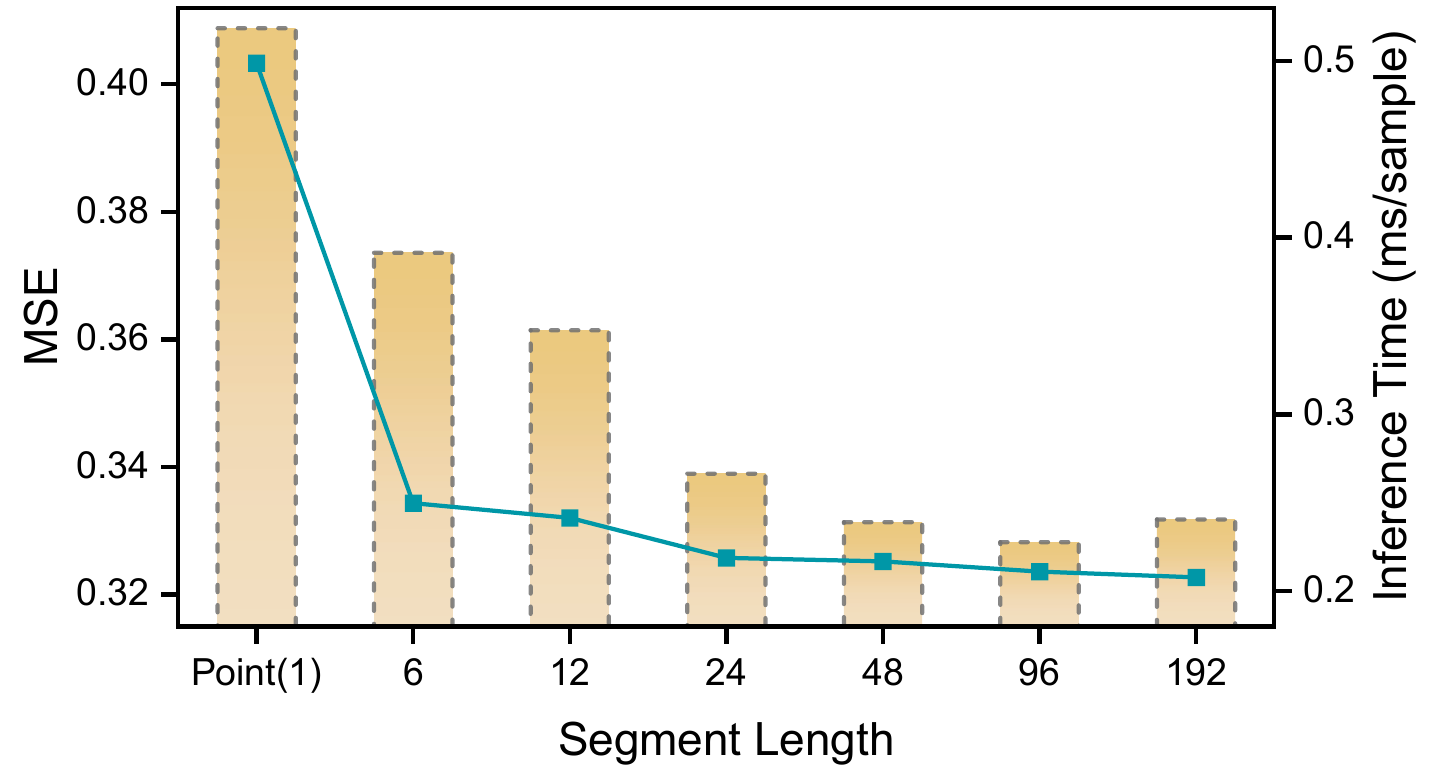}
  \caption{The forecast error (bar plot) and the inference time (line plot) of SegRNN with different segment length on ETTm1 dataset. Both look-back and horizon are 192.}
  \label{fig6}
\end{figure}

Figure \ref{fig6} illustrates the performance differences between segment-wise and point-wise iterations. It is essential to note that the segment length directly determines the number of iterations, and when the segment length \(w=1\), segment-wise iterations degenerates into point-wise iterations. The following observations can be made:

\begin{enumerate}
    \item As the segment length increases (i.e., the number of iterations decreases), the forecast error consistently decreases. However, when the segment length equals the look-back length, the model degenerates into a multi-layer perceptron, leading to an increase in prediction error.
    \item With the continuous increase of the segment length, the inference time steadily decreases.
\end{enumerate}

These findings indicate that a relatively large yet appropriate segment length (i.e., minimizing the number of iterations) significantly improves the performance of the RNN method in LTSF.

\paragraph{PMF vs. RMF.}

\begin{figure}[htb]
  \centering
  \includegraphics[width=0.95\columnwidth]{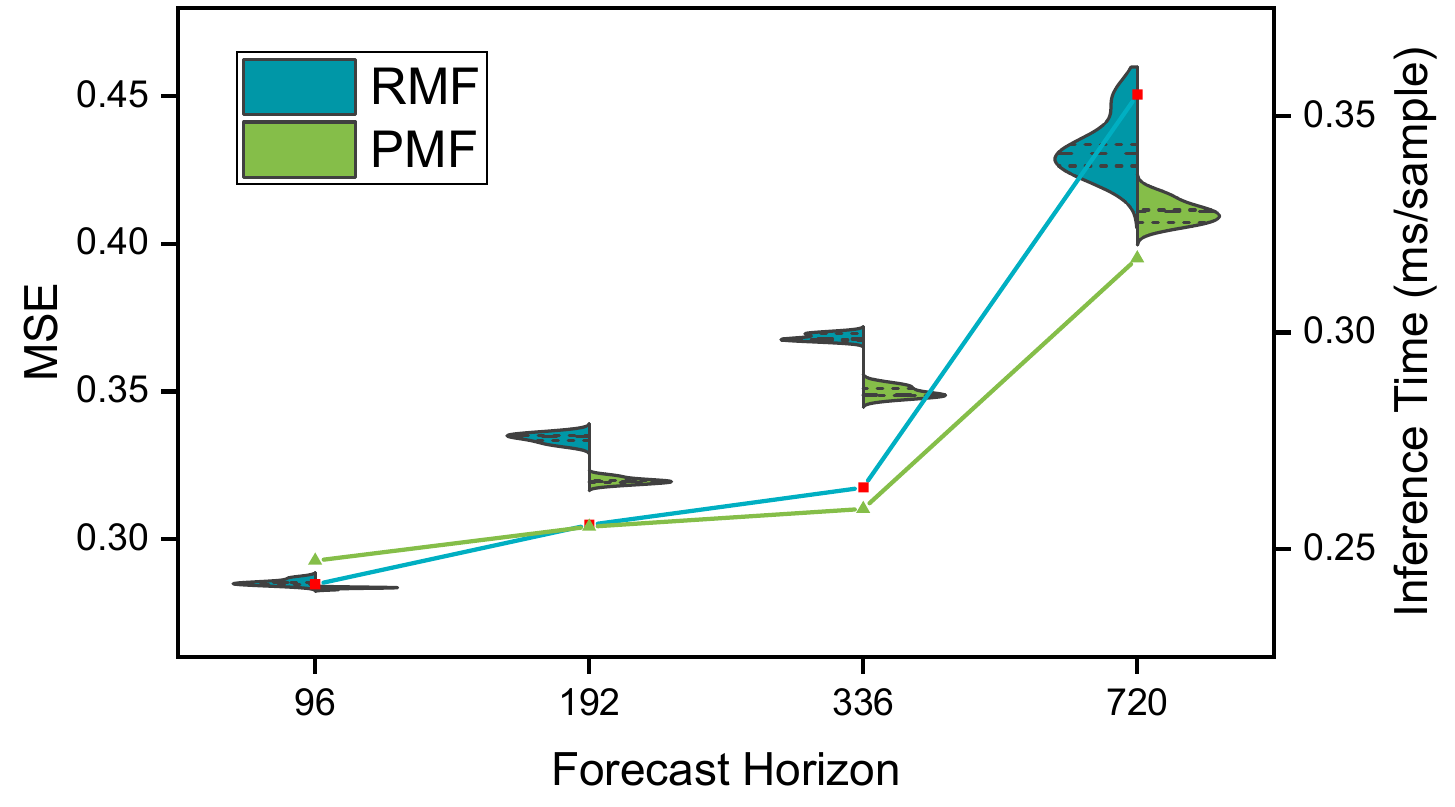}
  \caption{The forecast error distribution (violin plot) and the inference time (line plot) of SegRNN with different decoding strategy (i.e. RMF and PMF) on ETTm1 dataset. The look-back is 720.}
  \label{fig7}
\end{figure}

Figure \ref{fig7} illustrates the performance disparities between RMF and PMF. Concerning the forecast error, PMF significantly outperforms RMF for various forecast horizons, exhibiting a more stable distribution. The advantage of PMF becomes increasingly evident as the forecast horizon increases.

As for inference time, when the forecast horizon \(H<192\), PMF is slightly slower than RMF. This is because the intra-sample parallel computation advantage of PMF is not fully manifested with fewer iterations; instead, it incurs additional overhead due to data replication in memory. However, when the forecast horizon \(H>192\), the larger number of iterations allows PMF to leverage its intra-sample parallel computation advantage, leading to improved hardware utilization and accelerated computation.

In conclusion, these results demonstrate that PMF significantly enhances the performance of RNNs in LTSF tasks compared to RMF.

\subsection{Model Analysis}

\paragraph{Impact of look-back length.}
\begin{figure}[htb]
  \centering
  \includegraphics[width=0.8\columnwidth]{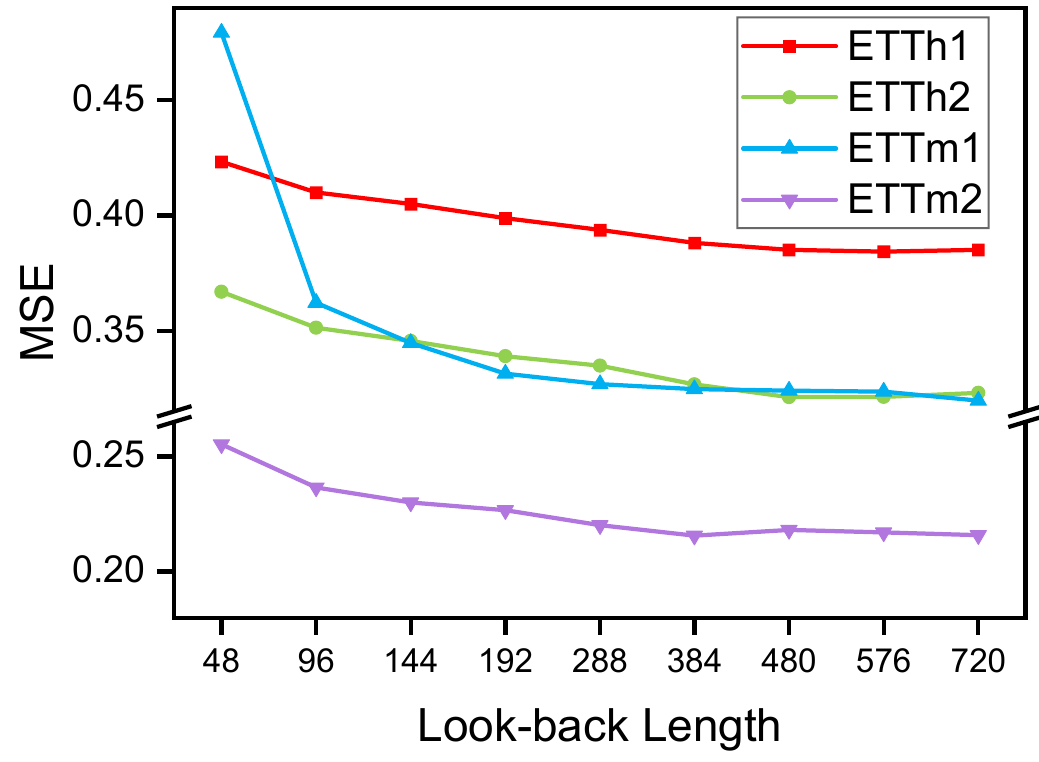}
  \caption{The forecast error of SegRNN with different look-back length on ETTm1 dataset. The horizon is 192.}
  \label{fig8}
\end{figure}
A powerful LTSF model typically performs better with a longer look-back context as it contains more trend and periodic information. The ability to leverage a longer look-back context directly reflects the model's capability to capture long-term dependencies. However, longer look-back contexts also increase the modeling complexity, and many Transformer-based models encounter difficulties when dealing with long look-back scenarios (i.e., \(L>96\)) \cite{dlinear}.

Regarding SegRNN, as illustrated in Figure \ref{fig8}, the forecast error consistently diminishes as the look-back length increases. Notably, SegRNN also exhibits commendable performance with relatively shorter look-backs, showcasing its robustness across diverse look-back lengths. This finding suggests that SegRNN not only excels in modeling long-term dependencies but also maintains its robustness in handling varying look-back requirements.

\paragraph{Impact of RNN variants.}
\begin{figure}[htb]
  \centering
  \includegraphics[width=0.8\columnwidth]{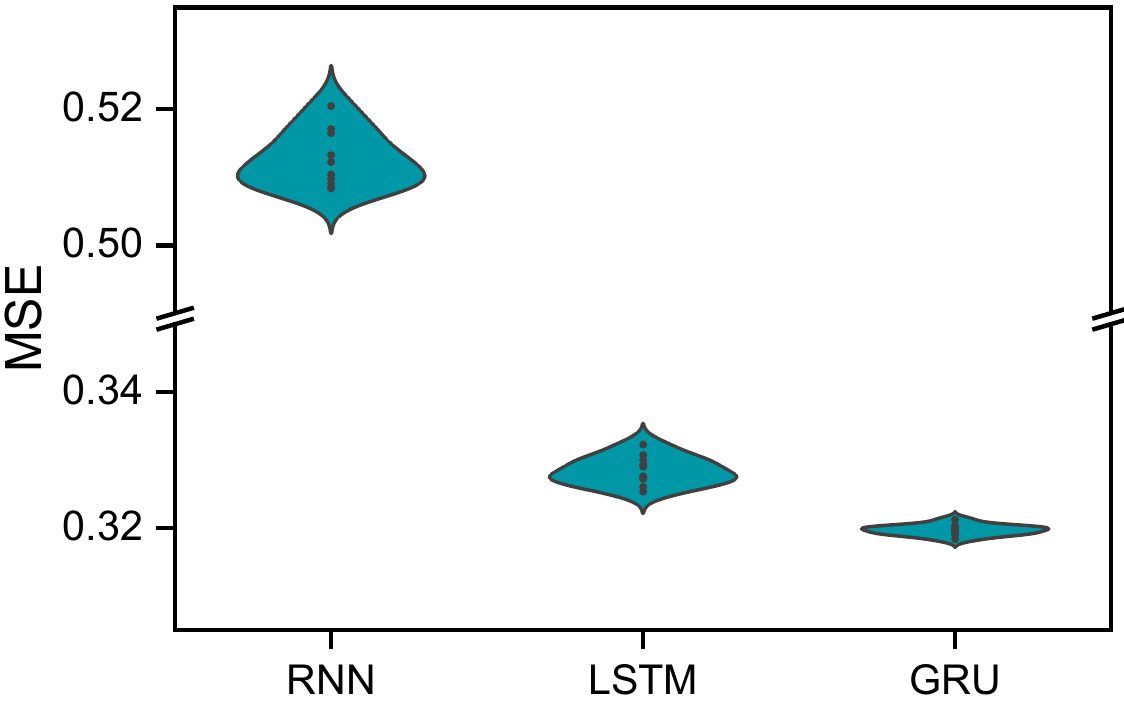}
  \caption{The forecast error of SegRNN with different RNN variants on ETTm1 dataset. The look-back is 720 and the horizon is 192.}
  \label{fig9}
\end{figure}
The proposed SegRNN exhibits versatility across various RNN types. Nevertheless, in practical implementations, it consistently achieves lower forecast errors and demonstrates greater stability when employed with the GRU variant, as depicted in Figure \ref{fig9}. This advantage is attributed to the integration of gating mechanisms in the GRU, which enhances its capacity to model long-term dependencies while retaining a simpler structure in comparison to LSTM. Consequently, SegRNN defaults to the adoption of GRU. However, should more potent RNN variants emerge in the future, SegRNN holds the potential to further bolster its robustness.

\paragraph{SegRNN vs. PatchTST.}
\begin{table}[htb]
\centering
\begin{adjustbox}{max width=0.85\columnwidth}
\begin{tabular}{@{}ccccc@{}}
\toprule
Metric & Datasets & PatchTST & SegRNN & Imp. \\ \midrule
\multirow{3}{*}{\begin{tabular}[c]{@{}c@{}}Training Time\\ (s/epoch)\end{tabular}} & ETTm1 & 94.9 & 29.7 & 69\% \\
 & Weather & 273.4 & 50.3 & 82\% \\
 & Electricity & 1.97k & 313.8 & 84\% \\ \midrule
\multirow{3}{*}{\begin{tabular}[c]{@{}c@{}}MACs\\ (MMac)\end{tabular}} & ETTm1 & 265.9 & 213.4 & 20\% \\
 & Weather & 797.8 & 640.2 & 20\% \\
 & Electricity & 12.2k & 9.79k & 20\% \\ \midrule
\multirow{3}{*}{\begin{tabular}[c]{@{}c@{}}Parameters\\ (M)\end{tabular}} & ETTm1 & 2.62 & 1.63 & 38\% \\
 & Weather & 2.62 & 1.63 & 38\% \\
 & Electricity & 2.62 & 1.71 & 35\% \\ \midrule
\multirow{3}{*}{\begin{tabular}[c]{@{}c@{}}Max Memory\\ (MB)\end{tabular}} & ETTm1 & 289 & 77 & 73\% \\
 & Weather & 780 & 124 & 84\% \\
 & Electricity & 11.3k & 1.23k & 89\% \\ \bottomrule
\end{tabular}
\end{adjustbox}
\caption{Comparison of performance metrics between SegRNN and PatchTST on ETTm1 dataset with a single NVIDIA T4 GPU. The look-back is 720 and the horizon is 192.}
\label{segrnn_vs_patchtst}
\end{table}

We conducted a comparison between SegRNN and the latest SOTA Transformer-based model, PatchTST, to showcase the runtime performance advantage of SegRNN. Table \ref{segrnn_vs_patchtst} reveals that compared to PatchTST, SegRNN demonstrates a reduction of over 78\% in average training time and a decrease of over 82\% in average maximum GPU memory consumption. This significant improvement in efficiency is particularly beneficial for practical model training and deployment.


\section{Conclusion}

In this paper, we introduce SegRNN, an innovative RNN-based model designed for Long-term Time Series Forecasting (LTSF). SegRNN incorporates two fundamental strategies: (i) the replacement of point-wise iterations with segment-wise iterations, and (ii) the substitution of Recurrent Multi-step Forecasting (RMF) with Parallel Multi-step Forecasting (PMF). The segment-wise iteration strategy significantly reduces the number of required recurrent iterations for extracting temporal features, thus addressing the challenge of effectively training RNNs on excessively long sequences. Moreover, the adoption of PMF further mitigates the issue of error accumulation inherent in traditional RMF methods. By adopting these innovative strategies, SegRNN not only outperforms the current SOTA models in terms of prediction accuracy but also yields substantial efficiency improvements, including a remarkable reduction of over 78\% in training time and memory usage. These compelling outcomes serve as robust evidence that RNNs remain potent in LTSF tasks, thereby encouraging further exploration and breakthroughs with more RNN methods in the future.

\appendix 
\section{Appendix}
In this section, we provide additional information, including more detailed experimental details, SegRNN's results in univariable long-term time series forecasting, convergence of different iteration schemes, and an analysis of the role of positional embeddings in PMF.
\subsection{Experimental details}
\paragraph{Datasets.}
We utilize the most popular multivariate datasets in Long-term Time Series Forecasting (LTSF), including:
\begin{itemize}
    \item ETTs\footnote{https://github.com/zhouhaoyi/ETDataset}: These datasets contain data collected from electricity transformers, including 7 indicators recorded from July 2016 to July 2018 in two regions of a province in China. ETTh1 and ETTh2 provide hourly-level data, while ETTm1 and ETTm2 offer data at a 15-minute granularity.
    \item Electricity: This dataset comprises hourly electricity consumption for 321 customers from 2012 to 2014.
    \item Traffic\footnote{https://pems.dot.ca.gov/}: Collected by the California Department of Transportation, this dataset provides hourly data describing road occupancy measured by 862 sensors on highways in the San Francisco Bay Area.
    \item Weather\footnote{https://www.bgc-jena.mpg.de/wetter/}: This dataset includes 21 meteorological indicators, such as temperature and humidity, recorded every 10 minutes throughout the year 2020.
\end{itemize}
We split all the datasets into training, validation, and test sets in chronological order. Following previous research \cite{Autoformer, dlinear, patchtst}, the ETT datasets are split into proportions of 6:2:2, while the other datasets are divided into proportions of 7:1:2.

\paragraph{Baselines.}
We selected various state-of-the-art (SOTA) deep learning methods in the LTSF domain as our baselines, including:

\begin{itemize}
\item PatchTST \cite{patchtst}: The most advanced Transformer-based method as of July 2023. It improves the performance of Transformers in LTSF by partitioning time series into patches and adopting a channel independent strategy. The patch technique not only inspired subsequent works such as TS-Mixer but also influenced the segment-wise iterations strategy proposed in this paper.
\item FEDformer \cite{fedformer}: This method combines Transformer with the seasonal-trend decomposition method and frequency-enhanced techniques, making it one of the classic Transformer-based approaches for time series forecasting.
\item Informer \cite{informer}: A pioneering method that introduced the prob-sparse self-attention mechanism, self-attention distilling, and the generative style decoder. It paved the way for Transformers in LTSF and discussed the challenges faced by RNNs in long-term modeling, which led to the rise of Transformer methods and the decline of RNNs in the LTSF domain.
\item DLinear \cite{dlinear}: An influential work that defeated the dominant Transformer-based methods, including FEDformer and Informer, by using trend decomposition and a single-layer linear approach. It questioned the necessity of attention mechanisms in modeling time series tasks. Additionally, it indirectly proposed the channel independence technique and brought attention to the importance of long look-back, inspiring subsequent research.
\item TiDE \cite{tide}: Clearly inspired by DLinear, TiDE inherits the channel independence technique and upgrades the linear layer to a multi-layer perceptron capable of modeling nonlinear dependencies. TiDE achieves comparable performance to PatchTST and is currently one of the most advanced MLP methods.
\item MICN \cite{micn}: This method introduces the Multi-scale Isometric Convolution Network to capture both local and global features of time series simultaneously. It is one of the most advanced CNN-based methods as of the current state.
\item TimesNet \cite{wu2023timesnet}: This method extends the analysis of temporal variations into 2D space by transforming 1D time series into a set of 2D tensors based on multiple periods, providing a task-general foundation model for time series analysis.
\item DeepAR \cite{deepar}: A classic RNN-based time series forecasting model that utilizes a deep LSTM network for Probabilistic forecasting. We modified it for long-term forecasting to use as a baseline.
\item GRU \cite{gru}: One of the powerful variants of RNN that mitigates the vanishing/exploding gradient problem. We employed it for long-term forecasting as the most basic RNN baseline.
\end{itemize}

For the results in Table \ref{main_result}, the data for PatchTST, FEDformer, and Informer are sourced from the PatchTST's official publication, while the data for DLinear, TiDE, MICN, and TimesNet are from their respective official papers. We implemented DeepAR and GRU ourselves, using a look-back of 720.

\paragraph{Configuration.}
\begin{table}[htb]
\centering
\begin{adjustbox}{max width=\columnwidth}
\begin{tabular}{@{}lccccccc@{}}
\toprule
Datasets & l\_back & s\_len & d\_model & channel & dropout & b\_size & l\_rate \\ \midrule
ETTh1 & 720 & 48 & 512 & True & 0.5 & 256 & 0.001 \\
ETTh2 & 720 & 48 & 512 & True & 0.5 & 256 & 0.0002 \\
ETTm1 & 720 & 48 & 512 & True & 0.5 & 256 & 0.0002 \\
ETTm2 & 720 & 48 & 512 & True & 0.5 & 256 & 0.0001 \\
Weather & 720 & 48 & 512 & True & 0.5 & 64 & 0.0001 \\
Electricity & 720 & 48 & 512 & True & 0.1 & 16 & 0.0005 \\
Traffic & 720 & 48 & 512 & False & 0.1 & 8 & 0.003 \\ \bottomrule
\end{tabular}
\end{adjustbox}
\caption{The complete configuration of SegRNN results in Table 2.}
\label{configuration}
\end{table}
\begin{table*}[t]
\centering
\begin{adjustbox}{max width=0.9\textwidth}
\begin{tabular}{@{}cccccccccccccccc@{}}
\toprule
\multicolumn{2}{c}{Models} & \multicolumn{2}{c}{\begin{tabular}[c]{@{}c@{}}SegRNN\\ (ours)\end{tabular}} & \multicolumn{2}{c}{\begin{tabular}[c]{@{}c@{}}PatchTST\\ (2023)\end{tabular}} & \multicolumn{2}{c}{\begin{tabular}[c]{@{}c@{}}Dlinear\\ (2023)\end{tabular}} & \multicolumn{2}{c}{\begin{tabular}[c]{@{}c@{}}MICN\\ (2023)\end{tabular}} & \multicolumn{2}{c}{\begin{tabular}[c]{@{}c@{}}FEDformer\\ (2022)\end{tabular}} & \multicolumn{2}{c}{\begin{tabular}[c]{@{}c@{}}Autoformer\\ (2021)\end{tabular}} & \multicolumn{2}{c}{\begin{tabular}[c]{@{}c@{}}Informer\\ (2021)\end{tabular}} \\ \midrule
\multicolumn{2}{c}{Metric} & MSE & MAE & MSE & MAE & MSE & MAE & MSE & MAE & MSE & MAE & MSE & MAE & MSE & MAE \\ \midrule
\multicolumn{1}{c|}{\multirow{4}{*}{ETTh1}} & \multicolumn{1}{c|}{96} & \textbf{0.053} & \textbf{0.18} & 0.059 & 0.189 & \underline{0.056} & \textbf{0.18} & 0.058 & \underline{0.186} & 0.079 & 0.215 & 0.071 & 0.206 & 0.193 & 0.377 \\
\multicolumn{1}{c|}{} & \multicolumn{1}{c|}{192} & \textbf{0.068} & \underline{0.208} & 0.074 & 0.215 & \underline{0.071} & \textbf{0.204} & 0.079 & 0.21 & 0.104 & 0.245 & 0.114 & 0.262 & 0.217 & 0.395 \\
\multicolumn{1}{c|}{} & \multicolumn{1}{c|}{336} & \textbf{0.073} & \textbf{0.215} & \underline{0.076} & \underline{0.22} & 0.098 & 0.244 & 0.092 & 0.237 & 0.119 & 0.270 & 0.107 & 0.258 & 0.202 & 0.381 \\
\multicolumn{1}{c|}{} & \multicolumn{1}{c|}{720} & \textbf{0.085} & \textbf{0.233} & \underline{0.087} & \underline{0.236} & 0.189 & 0.359 & 0.138 & 0.298 & 0.142 & 0.299 & 0.126 & 0.283 & 0.183 & 0.355 \\ \midrule
\multicolumn{1}{c|}{\multirow{4}{*}{ETTh2}} & \multicolumn{1}{c|}{96} & \textbf{0.121} & \underline{0.272} & 0.131 & 0.284 & 0.131 & 0.279 & 0.155 & 0.3 & \underline{0.128} & \textbf{0.271} & 0.153 & 0.306 & 0.213 & 0.373 \\
\multicolumn{1}{c|}{} & \multicolumn{1}{c|}{192} & \textbf{0.158} & \underline{0.317} & 0.171 & 0.329 & 0.176 & 0.329 & \underline{0.169} & \textbf{0.316} & 0.185 & 0.330 & 0.204 & 0.351 & 0.227 & 0.387 \\
\multicolumn{1}{c|}{} & \multicolumn{1}{c|}{336} & \underline{0.18} & \underline{0.345} & \textbf{0.171} & \textbf{0.336} & 0.209 & 0.367 & 0.238 & 0.384 & 0.231 & 0.378 & 0.246 & 0.389 & 0.242 & 0.401 \\
\multicolumn{1}{c|}{} & \multicolumn{1}{c|}{720} & \textbf{0.205} & \textbf{0.365} & \underline{0.223} & \underline{0.38} & 0.276 & 0.426 & 0.447 & 0.561 & 0.278 & 0.420 & 0.268 & 0.409 & 0.291 & 0.439 \\ \midrule
\multicolumn{1}{c|}{\multirow{4}{*}{ETTm1}} & \multicolumn{1}{c|}{96} & \textbf{0.026} & \textbf{0.121} & \textbf{0.026} & \underline{0.123} & \underline{0.028} & \underline{0.123} & 0.033 & 0.134 & 0.033 & 0.140 & 0.056 & 0.183 & 0.109 & 0.277 \\
\multicolumn{1}{c|}{} & \multicolumn{1}{c|}{192} & \textbf{0.039} & \underline{0.152} & \underline{0.04} & \textbf{0.151} & 0.045 & 0.156 & 0.048 & 0.164 & 0.058 & 0.186 & 0.081 & 0.216 & 0.151 & 0.310 \\
\multicolumn{1}{c|}{} & \multicolumn{1}{c|}{336} & \textbf{0.052} & \underline{0.177} & \underline{0.053} & \textbf{0.174} & 0.061 & 0.182 & 0.079 & 0.21 & 0.084 & 0.231 & 0.076 & 0.218 & 0.427 & 0.591 \\
\multicolumn{1}{c|}{} & \multicolumn{1}{c|}{720} & 0.081 & 0.223 & \textbf{0.073} & \textbf{0.206} & \underline{0.08} & \underline{0.21} & 0.096 & 0.233 & 0.102 & 0.250 & 0.110 & 0.267 & 0.438 & 0.586 \\ \midrule
\multicolumn{1}{c|}{\multirow{4}{*}{ETTm2}} & \multicolumn{1}{c|}{96} & \textbf{0.059} & \textbf{0.174} & 0.065 & 0.187 & \underline{0.063} & 0.183 & \textbf{0.059} & \underline{0.176} & 0.067 & 0.198 & 0.065 & 0.189 & 0.088 & 0.225 \\
\multicolumn{1}{c|}{} & \multicolumn{1}{c|}{192} & \textbf{0.084} & \textbf{0.215} & 0.093 & 0.231 & \underline{0.092} & \underline{0.227} & 0.1 & 0.234 & 0.102 & 0.245 & 0.118 & 0.256 & 0.132 & 0.283 \\
\multicolumn{1}{c|}{} & \multicolumn{1}{c|}{336} & \textbf{0.108} & \textbf{0.25} & 0.121 & 0.266 & \underline{0.119} & \underline{0.261} & 0.153 & 0.301 & 0.130 & 0.279 & 0.154 & 0.305 & 0.180 & 0.336 \\
\multicolumn{1}{c|}{} & \multicolumn{1}{c|}{720} & \textbf{0.153} & \textbf{0.304} & \underline{0.172} & 0.322 & 0.175 & \underline{0.32} & 0.21 & 0.354 & 0.178 & 0.325 & 0.182 & 0.335 & 0.300 & 0.435 \\ \midrule
\multicolumn{2}{c|}{Count} & \multicolumn{2}{c}{\textbf{30}} & \multicolumn{2}{c}{\underline{17}} & \multicolumn{2}{c}{14} & \multicolumn{2}{c}{5} & \multicolumn{2}{c}{2} & \multicolumn{2}{c}{0} & \multicolumn{2}{c}{0} \\ \bottomrule
\end{tabular}
\end{adjustbox}
\caption{Univariate long-term time series forecasting results. The forecast horizon \(H \in \{96, 192, 336, 720\}\) is set for all datasets. The reported SegRNN results are averaged over 5 runs. The best results are highlighted in \textbf{bold} and the second best are \underline{underlined}. The \textit{Count} row counts the total number of times each method obtained the best or second results.} 
\label{univariate}
\end{table*}

We employed the Adam optimizer \cite{adam} to train the models for 30 epochs, with a learning rate decay of 0.8 after the initial 3 epochs. Early stopping was implemented with a patience of 10. The specific parameters used for SegRNN on different datasets are presented in Table \ref{configuration}. The meanings of each parameter in the table are as follows:

\begin{itemize}
    \item l\_back: The length of the historical look-back window.
    \item s\_len: The window length for dividing the original sequence into segments.
    \item d\_model: The dimensionality of the hidden variables in the RNN layer.
    \item channel: Whether to enable channel positional embeddings.
    \item dropout: The dropout rate.
    \item b\_size: The batch size used for training.
    \item l\_rate: The initial learning rate used in the optimization process.
\end{itemize}

\subsection{Univariate Forecasting Results}

The univariate long-term time series forecasting results of SegRNN and other baselines on the full ETT benchmarks are presented in Table \ref{univariate}. The channel position encoding is disabled in univariate scenarios. Remarkably, SegRNN achieved a top-two position in 30 out of 32 metrics across all scenarios, including 23 first-place rankings, signifying its significant superiority over other baselines. These results further demonstrate the effectiveness of SegRNN and reaffirm the competitiveness of RNN methods in LTSF, whether in multivariate or univariate scenarios.

\subsection{Convergence of Different Iteration Schemes}
\begin{figure}[htb]
  \centering
  \includegraphics[width=0.95\columnwidth]{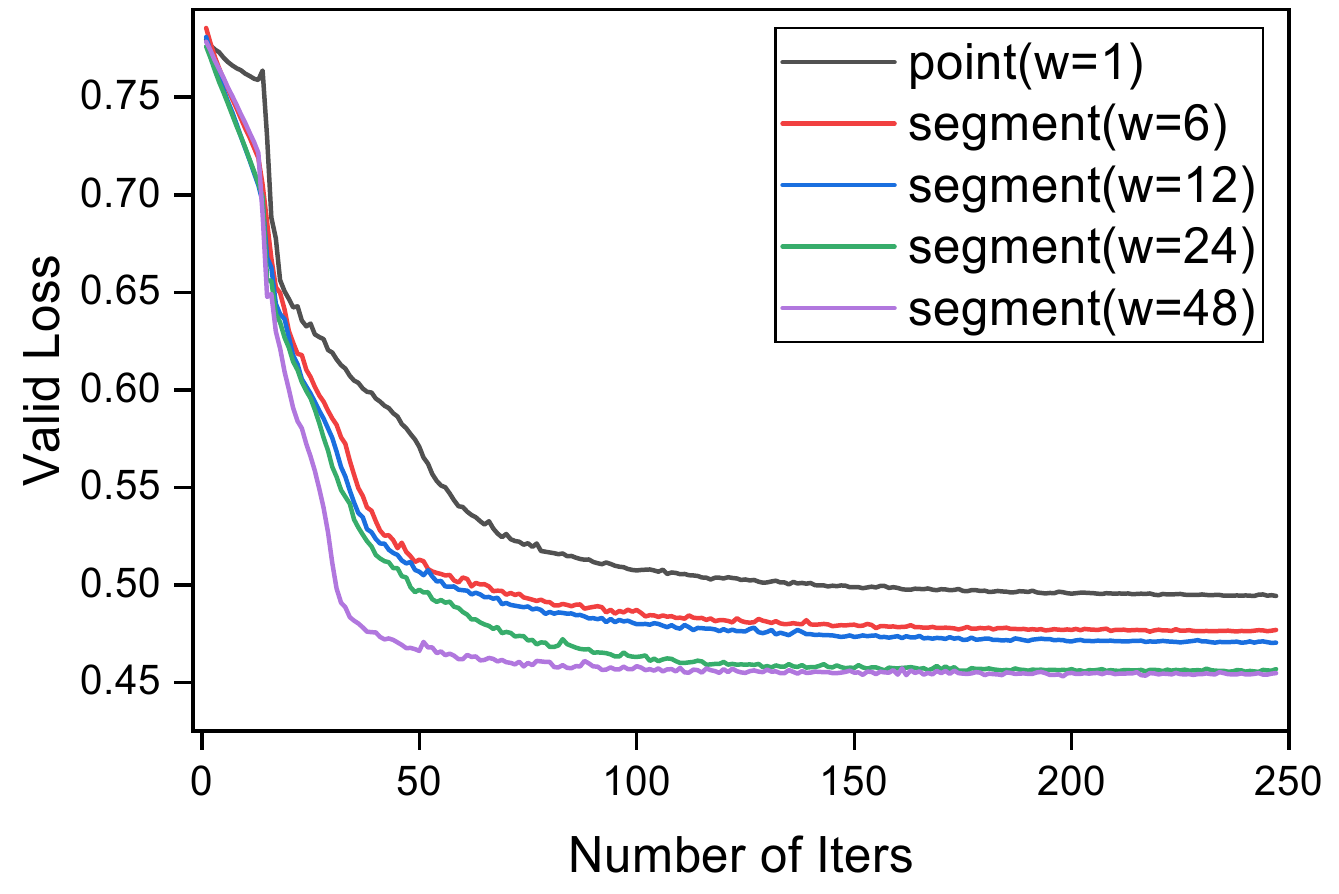}
  \caption{The convergence curve of point-wise iterations and segment-wise iterations on ETTm1 dataset.}
  \label{fig10}
\end{figure}

Figure \ref{fig10} illustrates the convergence curve of point-wise iterations and segment-wise iterations. It is evident that segment-wise iterations have considerably enhanced both the final convergence error and the convergence speed when compared to point-wise iterations. Furthermore, as the segment length increases (indicating a decrease in the number of segment-wise iterations), this improvement becomes more pronounced. Conclusionly, reducing the number of iterations as much as possible facilitates the convergence of the RNN and consequently enhances its performance.

Translated with www.DeepL.com/Translator (free version)

\subsection{Effect of Positional Embeddings}
\begin{table}[htb]
\centering
\begin{adjustbox}{max width=\columnwidth}
\begin{tabular}{lcccc}
\hline
Datasets & RP+CP & RP & CP & None \\ \hline
ETTm1 & \textbf{0.339} & \underline{0.340} & 0.683 & 0.691 \\
ETTm2 & \textbf{0.241} & \underline{0.245} & 0.274 & 0.278 \\
Electricity & \textbf{0.161} & \underline{0.162} & 0.210 & 0.209 \\
Traffic & \underline{0.604} & \textbf{0.596} & 0.806 & 0.778 \\
Weather & \textbf{0.219} & \underline{0.228} & 0.243 & 0.250 \\ \hline
Avg. & \textbf{0.312} & \underline{0.314} & 0.443 & 0.441 \\ \hline
\end{tabular}
\end{adjustbox}
\caption{Ablation study of Relative Position (RP) encoding and Channel Position (CP) encoding in Positional Embeddings. The results are the average Mean Squared Error (MSE) for forecast horizons \(H \in \{96, 192, 336, 720\}.\)}
\label{pe}
\end{table}
Table \ref{pe} presents the results of the ablation study on the effect of Positional Embeddings (PE). It can be observed that the Relative Position (RP) encoding plays a crucial role in PE. Compared to the original scheme without RP (None), the MSE is reduced by 28.8\%. This is because, for Parallel Multi-step Forecasting (PMF), the sequential order between segments is lost, and thus, it is essential to provide the model with relative positional information. As for the Channel Position (CP) encoding, it also contributes to some performance improvements. CP compensates for the missing relationship information between variables in the Channel Independent (CI) strategy, which has been extensively studied in STID \cite{STID}. However, in the case of Traffic, it might be an exception due to having over 800 variables. Modeling overly complex variable information solely through simple CP encoding might not be sufficient. Therefore, in the future, finding more reasonable ways to model complex multivariate relationships in time series forecasting will be a challenging yet promising direction.

\bibliography{myref}
\end{document}